\pdfoutput=1

\documentclass[11pt]{article}

\usepackage[final]{acl}

\usepackage{times}
\usepackage{latexsym}
 \usepackage{xcolor}
 
\usepackage[T1]{fontenc}

\usepackage[utf8]{inputenc}

\usepackage{microtype}

\usepackage{inconsolata}

\usepackage{bm}
\usepackage{tikzducks}
\usepackage{graphicx}
\usepackage[most]{tcolorbox}
\usepackage{colortbl}
\usepackage{pifont}
\usepackage{enumitem}
\definecolor{Gray}{gray}{0.9}

\usepackage{listings}
\definecolor{codeblue}{rgb}{0.25,0.5,0.5}
\definecolor{codekw}{rgb}{0.85, 0.18, 0.50}
\lstdefinestyle{mystyle}{
    backgroundcolor=\color{white},
    basicstyle=\fontsize{7.5pt}{7.5pt}\ttfamily\selectfont,
    columns=fullflexible,
    breaklines=true,
    captionpos=b,
    commentstyle=\fontsize{7.5pt}{7.5pt}\color{codeblue},
    keywordstyle=\fontsize{7.5pt}{7.5pt}\color{codekw},
}
\usepackage{graphicx}
\usepackage{amsmath, amssymb, bm}
\usepackage{algorithm}
\usepackage{multirow}
\usepackage{graphicx}
\usepackage{booktabs}
\usepackage{multicol}
\usepackage{hyperref}
\usepackage{amsmath}
\usepackage{caption}
\usepackage{enumitem}
\usepackage{wrapfig}
\usepackage{listings}
\usepackage{pifont}
\usepackage{array}
\usepackage{placeins}
\usepackage{float}
\definecolor{codegreen}{rgb}{0.0, 0.411, 0.243}
\definecolor{codered}{rgb}{0.89, 0.26, 0.20}
\usepackage{hyperref}
\definecolor{dartgreen}{HTML}{00693e}
\definecolor{refcolor}{HTML}{00693e}
\hypersetup{
    colorlinks=true,
    linkcolor=refcolor,
    citecolor=refcolor,
    filecolor=magenta,      
    urlcolor=refcolor,
    }

\title{Growing Through Experience:\\Scaling Episodic Grounding in Language Models}
\author{
  Chunhui Zhang$^{*1}$ \quad Sirui (Elsie) Wang$^{*1}$ \quad Zhongyu Ouyang$^1$ \\
  \textbf{Xiangchi Yuan}$^2$ \quad \textbf{Soroush Vosoughi}$^{1}$ \\
  $^1$Department of Computer Science, Dartmouth College \\
  $^2$School of Computer Science, Georgia Institute of Technology \\
  \small{\texttt{\{chunhui.zhang.gr, elsie.wang.gr, zhongyu.ouyang.gr, soroush.vosoughi\}@dartmouth.edu}} \\
  \small{\texttt{xyuan300@gatech.edu}}
}

\begin{document}
\maketitle
\begingroup
\renewcommand\thefootnote{}\footnotetext{
\noindent $*$: Equal contribution. Correspondence to \texttt{soroush.vosoughi@dartmouth.edu}
}
\endgroup
\begin{abstract}
Language models (LMs) require robust episodic grounding—the capacity to learn from and apply past experiences—to excel at physical planning tasks. Current episodic grounding approaches struggle with scalability and integration, limiting their effectiveness, especially for medium-sized LMs (7B parameters). While larger LMs (70–405B parameters) possess superior hierarchical representations and extensive pre-trained knowledge, they encounter a fundamental \textbf{scale paradox}: despite their advanced abstraction capabilities, they lack efficient mechanisms to leverage experience streams. We propose a scalable weak-to-strong episodic learning framework that effectively transfers episodic behaviors from smaller to larger LMs. This framework integrates Monte Carlo tree search for structured experience collection with a novel distillation method, preserving the inherent LM capabilities while embedding episodic memory. Experiments demonstrate our method surpasses state-of-the-art proprietary LMs by 3.45\% across diverse planning and question-answering tasks. Layer-wise probing further indicates significant improvements in task alignment, especially within deeper LM layers, highlighting stable generalization even for previously unseen scenarios with increased planning complexity—conditions where baseline methods degrade markedly.
\end{abstract}

\section{Introduction}
\label{sec:intro}
\begin{figure}
    \centering
    \includegraphics[width=0.485\textwidth]{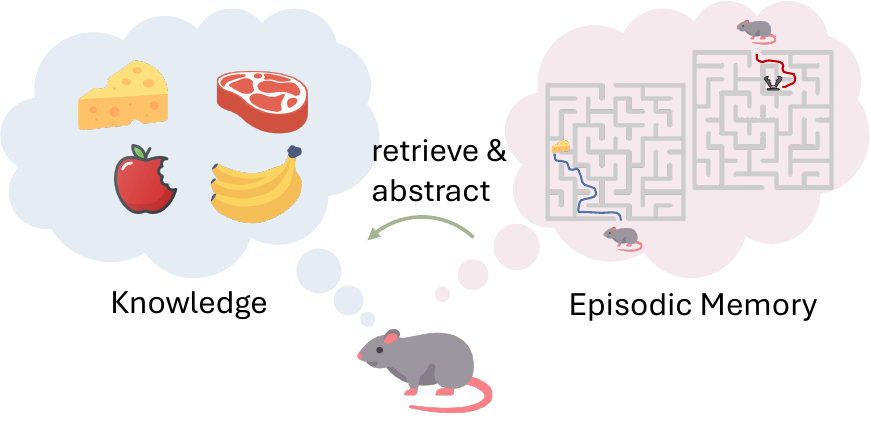}
    \caption{Brain cognition integrates \textbf{episodic memory} (specific events collected from explorations) into \textbf{generalized knowledge} through abstraction and retrieval. This hierarchical process parallels how LMs ground episodic experiences into context-aware planning and decision-making.}
    \label{fig:intro}
\end{figure}

Language models (LMs) have emerged with massive abilities to conduct diverse generation tasks~\cite{brown2020language, wei2022emergent, singhal2023large, yang2025navajo, yang2025visibility}, however, they still struggle to effectively plan physical tasks due to their limited ability to ground decisions in past experiences, with current approaches showing significant performance degradation in complex physical planning (see Figure~\ref{fig:increasing-planning-complexity}). This challenge parallels the crucial role of episodic memory in brain cognition, where organisms rely on stored experiences to adapt to new environments and make context-aware decisions \cite{varela2017embodied}, as Figure \ref{fig:intro} shows. 

Conceptually inspired by the hierarchical neural memory systems of the neocortex, our method leverages structured episodic experience collection and abstraction mechanisms analogous to those utilized by natural intelligence to encode and retrieve detailed, context-specific experiences \cite{dickerson2010episodic, moscovitch2016episodic}, subsequently generalizing them into broader world knowledge \cite{tulving1983elements, murty2016episodic}. Although our framework does not explicitly mimic the precise neural architectures underlying biological cognition, it draws on these fundamental principles to facilitate effective planning and decision-making in dynamic environments \cite{bartlett1995remembering, baddeley1992working}.

Current approaches to physical planning with LMs reveal a critical scaling challenge: small-sized LMs (1.3B parameters) achieve only 54.76\% accuracy on planning tasks, significantly underperforming larger LMs' (405B) 74.34\% accuracy \cite{xiang2023language}. This gap stems from small-sized models' limited capacity for hierarchical representation and long-term contextual recall, making them unable to effectively encode and retrieve episodic experiences \cite{pmlr-v235-das24a}. Even advanced test-time scaling techniques fail to bridge this gap, as small LMs still experience over 30\% performance degradation on complex planning sequences (see Figure~\ref{fig:increasing-planning-complexity}). \textit{Thus, there is no shortcut to enabling episodic grounding without leveraging the large pre-trained capabilities.}

While larger LMs (70-405B) show promise through their deep hierarchical architectures and extensive pre-trained knowledge, current solutions lack efficiency to integrate episodic experiences into their existing capabilities \cite{ichter2022do, pmlr-v202-driess23a}. This mirrors the cognitive challenges faced by individuals with impaired episodic memory, who struggle to adapt based on past experiences \cite{nuxoll2004cognitive}.
Therefore, we identify a fundamental \textbf{``scale paradox''} in episodic grounding: the largest models possess the necessary abstraction capacity for effective episodic reasoning but lack accessible fine-tuning paths at scale to incorporate experiences. Smaller models can be more easily trained on episodic data but lack the representational capacity to fully leverage these experiences. This architectural asymmetry, if left unresolved, prevents large LMs from accessing the experience streams that would enable their true potential.

To address this, we design weak-to-strong episodic grounding that combines Monte Carlo Tree Search (MCTS)-based experience collection with preference optimization, explicitly leveraging the pre-trained capabilities of large LMs. Our approach scales episodic grounding in a way that respects both the architectural and behavioral constraints of the models involved. It includes: structured episodic experience collection using MCTS, which generates both successful and failed exploration trajectories; a weak-to-strong distillation that leverages behavior ratios between post-trained and naive small models to guide larger models; and preference-based optimization that incorporates failed attempts as negative examples, enabling learning from episodic experiences.
By building on the pre-trained capabilities of large LMs, our framework overcomes the limitations of small-sized models and episodic memory scaling in physical planning tasks. This two-stage framework bridges the \textit{architectural asymmetry} with structured behavioral alignment, allowing large LMs to inherit task-specific grounding while preserving their generalist capacity.

This work shows the scaling effect of using pre-trained capabilities for episodic grounding in physical planning tasks. We achieve significant advances in physical planning capabilities: \underline{First,} our framework enables 70B and 405B LMs to resiliently maintain proper accuracy even in complex long-step planning sequences, while baseline methods show severe degradation beyond four steps. This scaling behavior demonstrates the effectiveness of our behavior ratio-based distillation in preserving planning capabilities across increasing task complexity. \underline{Second,} in comprehensive evaluations across physical planning and QA tasks, our approach outperforms advanced proprietary LLMs by 3.45\%, highlighting the value of learning from both successful and failed episodic experiences. \underline{Third,} through layer-wise probing analysis, we show that later model layers achieve up to 90\% accuracy in episodic reasoning tasks, providing empirical evidence for the emergence of hierarchical processing similar to human neocortical function. These results provide a practical framework for developing more capable AI systems that can effectively learn from and apply past experiences in complex, dynamic environments by leveraging the scaled capabilities of large LMs.

\section{Related Work}
\label{sec:related}
\paragraph{Embodied AI and Physical Simulators}
Physical simulators serve as virtual testbeds for training and evaluating AI models before real-world deployment. These simulators replicate real-world environments, enabling agents to interact with environments. Notable examples include VirtualHome~\cite{puig2018virtualhome, puig2021watchandhelp, xiang2023language, jin-etal-2024-mmtom}, a 3D household environment built using the Unity3D game engine, and ProcTHOR~\cite{procthor}, which procedurally generated scenes with rich object attributes and interaction types. Other simulators \cite{misra-etal-2018-mapping, yang2024learning} provide diverse environments for training embodied agents. More open-ended environments like MineCraft~\cite{fan2022minedojo, wang2024voyager} offer large-scale task hierarchies and objectives, making them particularly challenging for AI models. 
In our study, we structure the episodic experiences sampled from physical simulators, then shape them as preference data for distinguishing between positive and negative experiences based on goal satisfaction, to elicit the grounding ability of scaled LM.

\paragraph{Grounding in LMs}
Grounding LMs in multimodal or physical environments is a critical step toward enabling them to perform planning tasks that require interaction with the world and understanding different modalities~\cite{zhang2022look, diao2024learning, jian2023bootstrapping, han2024infimm, jian2024expedited, liu2025modality, diao2025temporal}. Recent works have explored various strategies to achieve this goal. One line of research focuses on leveraging frozen LMs with specialized prompts or auxiliary modules. For instance, Zero-Shot Planner~\cite{huang2022language} prompts LMs to generate activity plans and translates them into admissible actions. Similarly, SayCan~\cite{ichter2022do} uses a learned affordance function to assist LMs in selecting valid actions, while DEPS~\cite{wang2023describe} incorporates a learned selector module to choose the most efficient path based on LM-generated descriptions and explanations. Another approach involves fine-tuning LMs for specific tasks in target environments. For example, \citet{li2022pre} fine-tune LMs using supervised learning for interactive decision-making, and \citet{carta2023grounding} ground LMs using online reinforcement learning. 
While these methods have shown promise, they are often limited to specific tasks or environments and do not fully leverage the scaling potential of larger LMs. Our study scales episodic grounding from medium to large LMs, investigating their hierarchical capacity through layer-wise probing analysis.

\section{Methodology}
\label{sec:method}
\begin{figure}[t]
    \centering
    \includegraphics[width=0.485\textwidth]{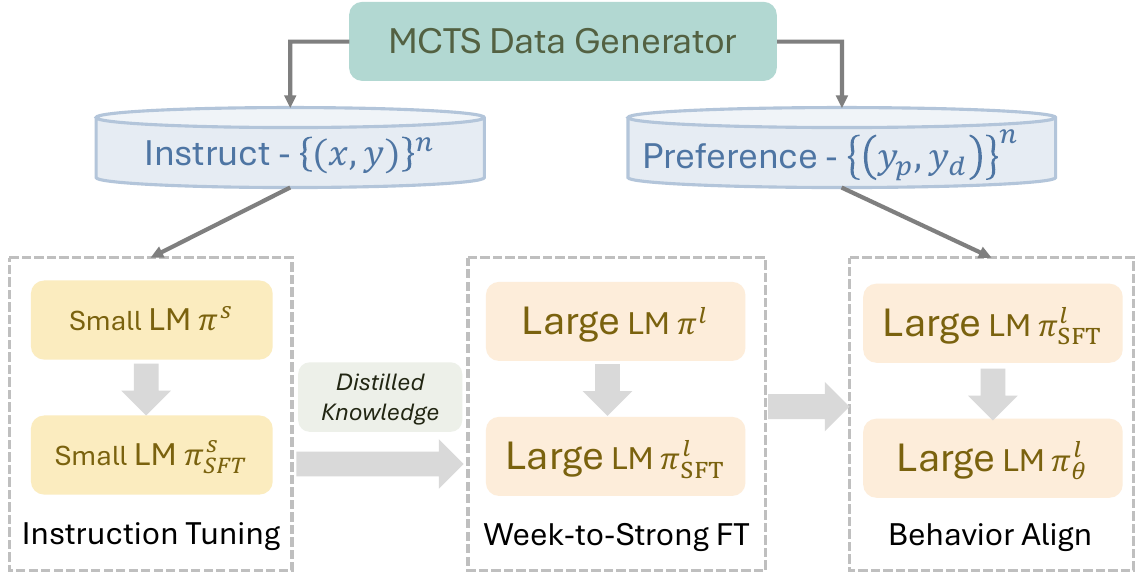}
    \caption{
    Overview of our weak-to-strong supervision framework for episodic grounding in LLM. Episodic experiences are collected through MCTS from a physical simulator and organized into instruction data $(x, y)$ and preference data $(y^+, y^-)$.
    }
    \label{fig:method}
\end{figure}

Our weak-to-strong supervision framework for episodic grounding in LLMs consists of two main stages: (1) experience collection using MCTS and (2) weak-to-strong distillation to transfer episodic behaviors from small to large LMs. Figure~\ref{fig:method} provides a pipeline overview.
\subsection{Experience Collection from MCTS}
We collect episodic data (instruction and preference) from a physical simulator (e.g., VirtualHome) using MCTS~\cite{xiang2023language}, which explores goal-oriented tasks and generates interaction histories. MCTS operates through four key steps:  
\textit{First}, in selection, the planner selects a promising node in the search tree using the Upper Confidence Bound:
   \begin{equation}
      UCT = Q(s, a) + C \cdot \sqrt{\frac{\log(N(s))}{N(s, a)}},
   \end{equation}
where $Q(s, a)$ is the estimated action $a$'s value in state $s$. $N(s)$ and $N(s, a)$ are visit counts.  
\textit{Second}, in expansion, a leaf node is expanded by adding a child node representing an unexplored action.  
\textit{Third}, in rollout, a sequence of actions is executed in the simulator, guided by a reward function (+2 for satisfying a goal predicate, -0.1 per timestep for irrelevant actions).  
\textit{Finally}, in backpropagation, rewards and state evaluations are propagated through the tree, updating $Q(s, a)$ and visit counts.  

Successful explorations (where all goal predicates are satisfied) are labeled as \textit{positive answers}, while failed explorations (where goals remain unmet) are labeled as \textit{negative answers}. To encourage the LM to generate compact and efficient plans, we also introduce redundant verbal styles as part of the negative answers. Specifically, during the simulation phase, we artificially create overly verbose or inefficient plans by adding unnecessary steps or repetitive actions. These redundant plans are then labeled as negative examples in the preference data, teaching the model to avoid such inefficiencies during plan generation. These paired experiences—compact and efficient plans as positive answers, and redundant or failed plans as negative answers—are incorporated into the LLM's post-training. This ensures that the model not only learns to recognize successful strategies but also avoids generating overly verbose or inefficient plans, significantly improving its ability to reason and act in goal-oriented contexts.

\subsection{Weak-to-Strong Episodic Grounding}
\paragraph{Training Small LMs on Episodic Experiences}  
To enable small LMs (under 8B parameters) to learn episodic grounding, we post-train them on episodic data collected via MCTS. These models are tasked with generating stepwise action sequences to achieve a given goal, starting from an initial state of relevant objects. Formally, we frame this as a sequence prediction problem, where the LM acts as a policy function $\pi$ that maps an input $\mathbf{x}$ (e.g., the initial condition) to an output sequence $\mathbf{y} = \{y_1, ..., y_M\}$ (e.g., the stepwise action sequence). The training objective is to maximize the likelihood of the correct action sequence, given the input and preceding actions:
\begin{equation}
\label{mix_training}
    \mathcal{L}_V = \sum\nolimits_{v \in V}\alpha_v\sum\nolimits_{m=1}^M \log \pi(y_m | \mathbf{y}_{<m}, \mathbf{x}),
\end{equation}
where $\mathcal{L}_V$ is the loss function for task set $V$, $\alpha_v$ is the weight for task $v$, $\mathbf{x}$ is the input, formatted as an instruction containing a task inquiry and in-context demonstrations, $\mathbf{y}_{<m}$ represents the sequence of actions generated up to step $m-1$.
By optimizing this objective, the small LM learns to predict the next action in a sequence, effectively internalizing the episodic grounding ability required for goal-oriented tasks. \textcolor{black}{This trained episodic grounding capability in smaller models provides a strong foundational behavior that larger LMs can subsequently inherit through a structured distillation process.} 

\paragraph{Episodic Distillation on Large LMs}
To scale episodic grounding to larger language models (LMs), we employ a weak-to-strong distillation approach that leverages the behavior shift observed in post-trained small LMs to adjust the output distribution of larger LMs in each decoding step, thereby enabling efficient knowledge transfer.
Formally, let $\pi^{\mathcal{E}}$ denote the policy distribution of a post-trained small LM, and $\pi^{\mathcal{N}}$ denote the policy distribution of a naive small LM. \textcolor{black}{This behavior shift, indicating learned episodic grounding,} is captured by the ratio $\frac{\pi^{\mathcal{E}}(y_m | \mathbf{y}_{<m}, \mathbf{x})}{\pi^{\mathcal{N}}(y_m | \mathbf{y}_{<m}, \mathbf{x})}$, which approximates the effect of post-training on the policy function. At each generation step $t$, the adjusted policy distribution for the large LM $\pi^{\mathcal{L}}$ is:
\begin{equation}
\begin{aligned}
\bar{\pi}(y_m | \mathbf{y}_{<m}, \mathbf{x}) 
    &= \frac{1}{\bar{Z}} \pi^{\mathcal{L}}(y_m | \mathbf{y}_{<m}, \mathbf{x}) \\
    &\quad \times \frac{\pi^{\mathcal{E}}(y_m | \mathbf{y}_{<m}, \mathbf{x})}
                      {\pi^{\mathcal{N}}(y_m | \mathbf{y}_{<m}, \mathbf{x})},
\end{aligned}
\label{eq:weak-to-strong}
\end{equation}
where $\bar{Z} = \sum_{a^t} \pi^{\mathcal{L}}(y_m) \frac{\pi^{\mathcal{E}}(y_m)}{\pi^{\mathcal{N}}(y_m )}$ is the normalization factor. \textcolor{black}{This formulation explicitly transfers episodic behaviors from smaller models, effectively guiding larger models without extensive computational overhead.} To further align the large LM with the post-trained behavior, we optimize the reverse KL-divergence loss:
\begin{equation}
\mathcal{L}_{\text{RKL}} = \mathbb{E}_{\mathbf{x}, \mathbf{y} \sim \bar{\pi}}\left[\sum_{m=1}^M \log \frac{\bar{\pi}(y_m | \mathbf{y}_{<m}, \mathbf{x})}{\pi^{\mathcal{L}}(y_m | \mathbf{y}_{<m}, \mathbf{x})}\right].
\end{equation}
Unlike forward KL-divergence, which encourages broad coverage of the target distribution, reverse KL-divergence induces mode-seeking behavior~\cite{gu2024minillm, wang2024beyond}. \textcolor{black}{This property ensures that larger LMs produce more precise, confident, and goal-aligned action sequences by leveraging the distilled episodic experiences.} 

\paragraph{Preference Optimization on MCTS Events}
While instruction tuning and distillation enable large LMs to learn from successful episodic experiences, these approaches often overlook the valuable information contained in failed attempts. This limitation can lead to overfitting and reduced generalization capability, as models lack exposure to counterexamples that illustrate suboptimal strategies~\cite{rafailov2024direct, wang2024beyond}. To address this challenge, we introduce a preference-based optimization framework that leverages both successful and failed explorations generated through MCTS, enabling episodic behavior learning.

Our approach extends Direct Preference Optimization (DPO)~\cite{rafailov2024direct} to the domain of physical planning by incorporating structured contrasts between successful and failed episodic experiences. Specifically, we construct preference pairs where positive samples ($y^+$) represent states achieving all goal predicates during MCTS exploration, while negative samples ($y^-$) correspond to states where goals remain unsatisfied. This binary classification framework enables the model to learn fine-grained distinctions between effective and ineffective strategies in goal-oriented contexts.

To optimize the large LM's decision-making capabilities, we formulate a modified DPO loss function that combines preference learning with distribution regularization:
\begin{equation}
\begin{aligned}
\mathcal{L}_{\text{DPO}} = & - \mathbb{E}_{(\mathbf{x}, y^+, y^-) \sim \mathcal{D}} \\
& \left[ \log \sigma\left( \beta \cdot \left( \log \pi(y^+ | \mathbf{x}) - \log \pi(y^- | \mathbf{x}) \right) \right) \right] \\
& + \lambda \cdot \mathbb{E}_{\mathbf{x}, y \sim \pi} \left[\log \frac{\pi(y | \mathbf{x})}{\pi_0(y | \mathbf{x})}\right].
\end{aligned}
\end{equation}
Here, $\beta$ serves as a temperature parameter controlling the sharpness of preference learning, while $\lambda$ weights the reverse KL divergence term that maintains proximity to the initial policy $\pi_0$. This formulation ensures stable preference learning while preserving the model's pre-trained capabilities~\cite{gu2024minillm}. The input $\mathbf{x}$ encodes both the goal specification and environmental state, allowing the model to learn context-dependent preferences across diverse scenarios.

\section{Experiments}
\label{sec:exp}
We investigate our weak-to-strong episodic grounding framework across three key dimensions: (1) effectiveness in physical planning and QA tasks compared to state-of-the-art baselines, (2) scalability benefits from 1.3B to 405B parameters, and (3) in-depth analysis of how episodic knowledge is processed across model layers. We further investigate the framework's resilience to increasing task complexity and reveal through probing analysis that our method approaches the upper bounds of model capabilities while maintaining the flexibility of next-token prediction.
\begin{table*}
\centering
\resizebox{1\textwidth}{!}{
\begin{tabular}{llccccccccccccccc}
\toprule
\multirow{2}{*}{LM} & \multirow{2}{*}{Config} & \multicolumn{6}{c}{Plan Generation} & \multicolumn{7}{c}{Question Answering} & \multirow{2}{*}{Avg.} \\
\cmidrule(lr){3-8} \cmidrule(lr){9-15}
  &  & VS & VU & CS & CU & Path & Avg. & HW & Neg. & Recog. & Inf. & Count. & Loc. & Avg. & \\  
\midrule
GPT-4o  & base & 52.67 & 49.35 & 47.54 & 46.22 & 81.23 & {55.40} & 85.37 & 84.31 & 95.60 & 84.85 & 78.43 & 74.21 & 83.80 & 70.89 \\  
\midrule
\multirow{2}{*}{GPT-Neo} & 1.3B-base & 21.25 & 17.64 & 16.86 & 17.05 & 30.80 & 20.72 & 70.11 & 38.27 & 69.22 & 56.49 & 22.68 & 22.50 & 46.55 & 34.81 \\
                        & 1.3B-ewc  & 49.70 & 49.27 & 46.88 & 42.34 & 85.91 & 54.82 & 72.41 & 41.98 & 85.43 & 66.03 & 28.87 & 33.50 & 54.70 & 54.76 \\
\midrule
\multirow{3}{*}{GPT-J} & 6B-base & 34.31 & 34.22 & 34.81 & 32.98 & 33.86 & 34.04 & 77.78 & 35.19 & 87.98 & 69.08 & 30.41 & 30.00 & 55.07 & 45.51 \\
                      & 6B-ft   & 47.98 & 47.86 & 47.59 & 44.43 & 46.25 & 46.82 & 51.34 & 33.33 & 71.41 & 70.99 & 16.49 & 22.50 & 44.68 & 45.75 \\
                      & 6B-ewc  & {51.23} & {49.58} & {48.94} & {45.60} & \textbf{98.67} & {58.80} & 85.44 & 39.51 & {88.52} & \underline{74.43} & 67.01 & 34.50 & 64.90 & 61.29 \\
\midrule
\multirow{2}{*}{OPT} & 13B-base & 36.00 & 29.34 & 31.92 & 36.98 & 33.49 & 33.95 & 81.61 & 43.21 & 89.07 & 67.94 & 20.01 & 37.00 & 56.14 & 45.04 \\
                     & 13B-ewc  & 50.15 & 45.11 & 49.87 & 47.93 & {96.28} & 57.07 & 84.29 & 40.21 & 91.44 & 70.61 & {62.37} & 33.00 & 63.32 & 60.58 \\
\midrule
\multirow{2}{*}{Llama1} & 13B-base & 41.77 & 38.78 & 40.33 & 41.73 & 38.82 & 40.29 & 81.99 & 43.21 & 90.53 & {74.05} & 29.38 & 28.50 & 57.28 & 48.78 \\
                       & 13B-ewc  & \underline{52.05} & 47.44 & \underline{51.00} & \underline{50.49} & \underline{96.99} & 59.99 & \underline{86.59} & 30.25 & {91.80} & 68.32 & \underline{79.38} & 79.00 & 72.56 & 66.28 \\
\midrule
\multirow{3}{*}{Llama2} & 13B-base & 39.97 & 38.81 & 39.50 & 37.55 & 67.46 & 44.26 & 83.25 & 52.00 & 89.25 & 69.50 & 32.54 & 31.40 & 59.66 & 51.96 \\
                       & 13B-ewc  & 42.43 & 43.50 & 43.10 & 46.32 & 78.38 & 50.75 & 82.49 & \underline{52.95} & 90.52 & 67.85 & 75.48 & 72.53 & 73.30 & 61.87 \\
                       & 70B-base & 46.77 & 40.68 & 39.34 & 34.18 & 78.64 & 47.92 & {85.82} & 33.95 & 92.35  & 69.47 & 71.65  & 80.50 & 72.29 & 61.21     \\
\midrule
\multirow{2}{*}{Llama3.1} 
                         & 70B-ours    & 48.43 & \underline{49.66} & 48.24 & 45.23 & 80.05 & \underline{61.73} & 84.23 & 38.33 & \textbf{93.54}  & 73.23 & 76.23  & \underline{81.32} & \underline{85.60} & \underline{73.34}   \\
                         & 405B-ours   & \textbf{56.37} & \textbf{55.82} & \textbf{56.05} & \textbf{56.32} & 86.98 & \textbf{62.71} & \textbf{86.60} & \textbf{85.42} & \underline{93.08} & \textbf{81.27} & \textbf{84.10} & \textbf{84.44} & \textbf{85.82} & \textbf{74.34} \\
\bottomrule
\end{tabular}
}
\caption{Results on various downstream evaluation tasks.  
\textbf{Plan Generation} tasks are evaluated using ROUGE-L for VS (Vanilla Seen), VU (Vanilla Unseen), CS (Confusing Seen), CU (Confusing Unseen), and Longest Common Subsequence for Path (Object Path Tracking).  
\textbf{Question answering} tasks are evaluated using accuracy for HW (Housework QA), Neg. (Negation QA), Recog. (Activity Recognition), Inf. (Activity Inference), Count. (Counting), and Loc. (Object Location QA). Base models represent standard pre-trained LMs without post-training, while ft (fine-tuned), ewc (\cite{xiang2023language}), and ours (our post-training pipeline) configurations illustrate performance gains from episodic post-training. We bold the best results and underline the best open-source baselines.}
\label{tab:main-results}
\end{table*}

\subsection{Setup}
\paragraph{Evaluation}
The evaluation datasets build upon the RobotHow knowledge base~\cite{xiang2023language}, implemented within the VirtualHome environment~\cite{puig2018virtualhome}, which provides a rich foundation for testing physical planning capabilities. The evaluation encompasses multiple task categories designed to assess different aspects of episodic grounding. For plan generation tasks, we utilize human-written plans from RobotHow to evaluate the model's ability to generate step-by-step instructions for household activities, with particular emphasis on unseen scenarios to test out-of-distribution generalization. The evaluation extends to plan-activity recognition tasks, where models must demonstrate understanding by inferring activity names from either comprehensive process descriptions or final states alone. These tasks assess the model's capacity to reason about cause-and-effect relationships and understand the physical progression of activities. 
\begin{table*}[t]
\centering
\resizebox{1\textwidth}{!}{
\begin{tabular}{llccccccccccccccc}
\toprule
\multirow{2}{*}{LM} & \multirow{2}{*}{Config} & \multicolumn{6}{c}{Plan Generation} & \multicolumn{7}{c}{Question Answering} & \multirow{2}{*}{Avg.} \\
\cmidrule(lr){3-8} \cmidrule(lr){9-15}
  &  & VS & VU & CS & CU & Path & Avg. & HW & Neg. & Recog. & Inf. & Count. & Loc. & Avg. & \\  
\midrule
\multirow{1}{*}{GPT-3.5} & base & \underline{40.57} & \underline{41.01} & \underline{40.41} & \underline{40.97} & \underline{59.53} & 44.50 & \underline{83.91} & \textbf{87.65} & \underline{95.05} & \underline{83.59} & \underline{66.49} & \underline{67.50} & 80.70 & \underline{64.24} \\  
\multirow{1}{*}{GPT-4o}  & base & \textbf{52.67} & \textbf{49.35} & \textbf{47.54} & \textbf{46.22} & \textbf{81.23} & \textbf{55.40} & \textbf{85.37} & 84.31 & \textbf{95.60} & \textbf{84.85} & \textbf{78.43} & \textbf{74.21} & \textbf{83.80} & \textbf{70.89} \\  
\midrule
\multirow{5}{*}{Llama3.1} & 8B-base     & 41.23 & 39.65 & 40.72 & 41.53 & 68.21 & 46.27 & 84.12 & 54.32 & 90.62 & 70.20 & 47.49 & 64.12 & 68.81 & 57.54 \\
& 8B-ft       & 53.42 & \underline{50.89} & \underline{52.36} & \underline{52.33} & \textbf{96.48} & 61.50 & \underline{86.23} & 53.70 & 92.52 & 75.65 & \underline{81.65} & 80.50 & 78.38 & 70.77 \\
& 70B-base    & \textbf{57.11} & 43.23 & 45.56 & 44.43 & 78.32 & 53.72 & 77.02 & \underline{76.50} & 89.06 & \textbf{82.08} & 78.35 & 70.57 & \underline{78.93} & 67.48 \\
& 70B-ours    & 48.43 & \underline{49.66} & 48.24 & 45.23 & \underline{80.05} & \underline{61.73} & 84.23 & 38.33 & \textbf{93.54} & 73.23 & 76.23 & \underline{81.32} & \underline{85.60} & \underline{73.34} \\
& 405B-ours   & \underline{56.37} & \textbf{55.82} & \textbf{56.05} & \textbf{56.32} & 86.98 & \textbf{62.71} & \textbf{86.60} & \textbf{85.42} & \underline{93.08} & \underline{81.27} & \textbf{84.10} & \textbf{84.44} & \textbf{85.82} & \textbf{74.34} \\
\bottomrule
\end{tabular}
}
\caption{Comparison of LMs (proprietary v.s. open-source) across different scales and specialization configurations.}
\label{tab:scaling-up-results}
\end{table*}

\paragraph{Baselines} We include established LMs of varying scales, including GPT-Neo-1.3B \cite{black2021gptneo}, GPT-J-6B \cite{wang2021gptj}, OPT-13B \cite{zhang2022opt}, and Llama series \cite{touvron2023llama1, touvron2023llama2, dubey2024llama3}. Each baseline undergoes specialization through different approaches—direct fine-tuning (ft), elastic weight consolidation (ewc)~\cite{xiang2023language}, and our proposed pipeline—ensuring a thorough assessment of episodic grounding capabilities across model architectures and training strategies.

\paragraph{Post-training} 
During instruction tuning, we train small-scale 8B LMs using a learning rate of $1 \times 10^{-3}$ over five epochs, establishing foundational episodic behaviors. The subsequent distillation phase employs a two-step approach: an initial bootstrapping epoch aligns the 70B parameter model with the small model's behavior, followed by a focused epoch of cross-entropy training on episodic data. For preference optimization, the KL penalty coefficient $\lambda = 1$ and learning rate is $1 \times 10^{-4}$, balancing the trade-off between preference learning and preservation of pre-trained capabilities. 
For larger-scale LMs (405B parameters), we implement an approximation strategy: rather than directly optimizing their weights, we capture behavioral changes from 70B LM trained by our solution and apply these adjustments to the 405B LM's output distribution, enabling scalable episodic grounding without prohibitive computational costs.

\subsection{Superior Performance Across Tasks}
According to Table~\ref{tab:main-results}, our solution achieves top performance across physical planning tasks, demonstrating the effectiveness of combining pretrained knowledge with structured episodic experiences. Compared to GPT-4o's overall performance 70.89, our approach achieves better score with 74.34, surpassing the best baseline by 3.45. This improvement emerges from the systematic integration of episodic knowledge across model scales, as evidenced by consistent performance gains in models ranging from 1.3B to 405B parameters when specialized on physical simulator data.

\textit{The framework's effectiveness is particularly pronounced in plan generation, where the integration of episodic experiences proves crucial for physical environment reasoning}. Our method achieves 62.71 in plan generation, significantly outperforming both GPT-4o (55.40) and the strongest open-source baseline, Llama1-13B-ewc (59.99). This superiority in planning tasks demonstrates the framework's scaled capability to encode and utilize structured episodic experiences for physical grounding.

\textit{In question-answering tasks, which traditionally favor models with extensive pretrained knowledge, our approach addresses a critical limitation of previous methods}. While larger LMs like Llama2-13B and Llama2-70B show strong baseline performance (59.66\% and 72.29\% respectively) compared to GPT-4o (83.80\%), our method achieves superior results by combining pretrained knowledge with episodic grounding. This synergy enables our model to achieve the highest QA accuracy of 85.82\%, while maintaining strong performance in plan generation.
The framework's success stems from its unique ability to \textit{leverage both extensive pretrained knowledge and structured episodic experiences through weak-to-strong training}. This dual advantage enables balanced performance across diverse task types, ultimately establishing a new state-of-the-art benchmark with 74.34\% overall accuracy—a 3.45\% improvement over baselines.

\subsection{Model Scale Effects on Episodic Learning}
The results, detailed in Table~\ref{tab:scaling-up-results}, detail the evidence for the scalability and stability of our approach.
\textit{Model scale demonstrates a clear positive correlation with episodic grounding performance, as evidenced by both parameter count and pretraining data volume.} This relationship manifests across different model families: GPT-4o (70.89) significantly outperforms GPT-3.5 (64.24), while within the Llama series, performance scales from Llama2-13B (51.96) to Llama3.1-8B (57.54). The latter's superior performance, despite fewer parameters, can be attributed to its expanded pretraining corpus of 15 trillion tokens compared to Llama2's 2 trillion. This scaling trend continues through the Llama3.1 family, progressing from 57.54 (8B) to 70.77 (8B-ft) and ultimately reaching 74.34 with the 405B model, suggesting further potential for improvement at even larger scales.
\textit{Post-training consistently enhances model performance across all scales, even in models with extensive pretraining.} This effect is particularly pronounced in Llama3.1-70B, where post-training improves performance from 67.48 to 73.34. The enhancement is especially significant in tasks involving physical planning and dynamic environments, where post-training helps align pretrained knowledge with specific task requirements. Our weak-to-strong distillation framework efficiently transfers these benefits to larger models through two approaches: for the 70B model, cross-entropy-based distillation from smaller, well-trained models achieves a 73.34 average score, while for the 405B model, we employ an innovative inference-time output adjustment method, reaching 74.34 and surpassing GPT-4o's 70.89 performance. This scalable approach demonstrates the framework's ability to effectively transfer episodic knowledge across model scales while maintaining computational efficiency.

\subsection{Probing LM's Full Potential}
We investigate whether our method better aligns the LM's internal representations with its full potential in downstream tasks. Standard evaluations, which focus on next-token prediction using the final layer's hidden state, often overlook the rich information encoded across all layers, especially in models post-trained with episodic knowledge from physical simulators.

\begin{figure}[t]
    \centering
    \includegraphics[width=0.48\textwidth]{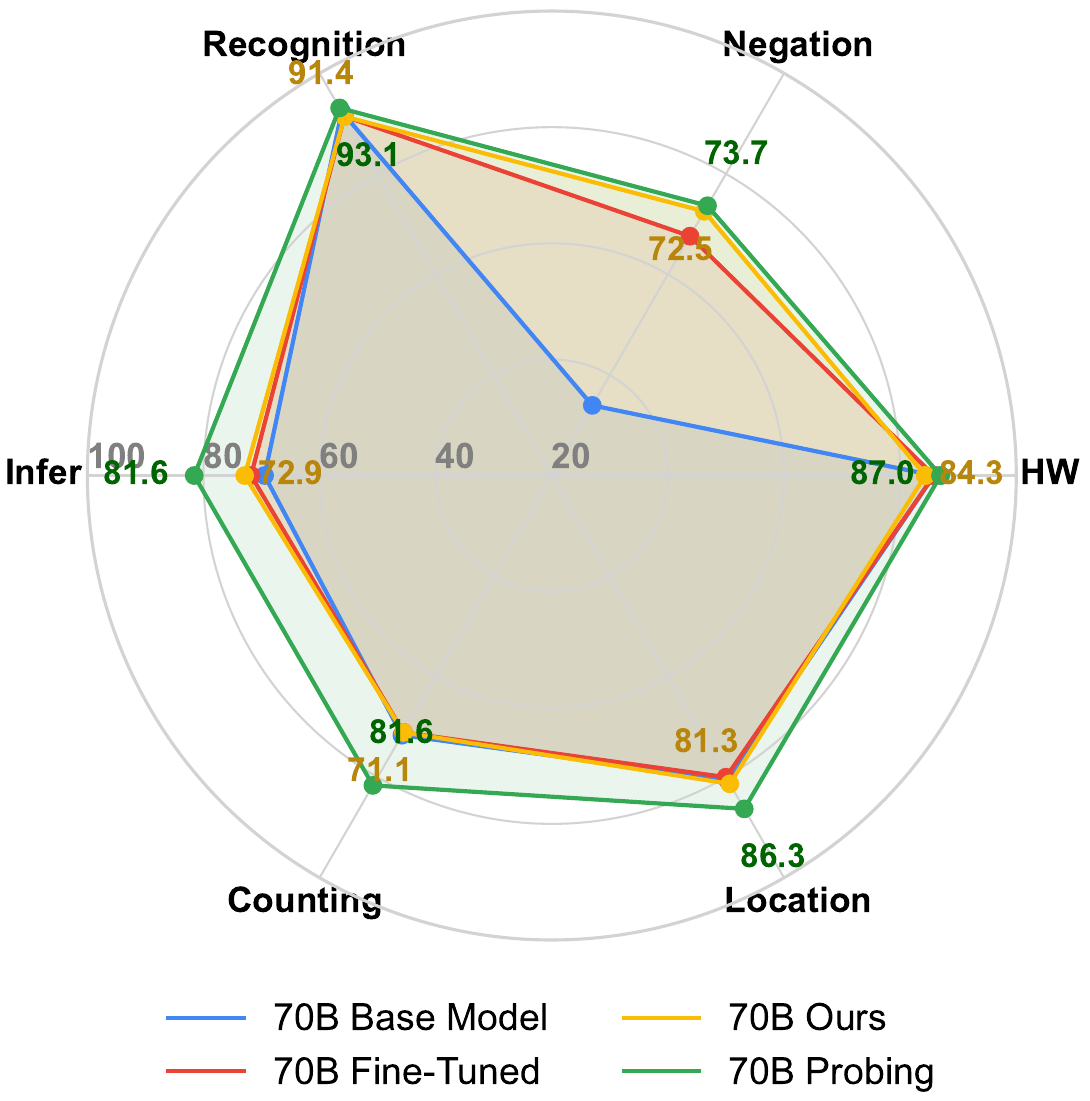}
    \caption{Accuracy comparison on six tasks for different configurations on Llama2. 70B-probing achieves the overall highest accuracy across six tasks and shows the boundary of LM representation's grounding potential.}
    \label{fig:70B-radar-qa}
\end{figure}
To touch an upper bound for the LM's capabilities, we adopt the probing for multi-choice question-answering~\cite{orgad2025llms,chetelat2025innerthoughts}. This approach leverages hidden states from all intermediate layers at the last temporal position, revealing the model's full potential beyond standard next-token prediction. As shown in Figure~\ref{fig:70B-radar-qa}, our method significantly narrows the gap between standard fine-tuning and the probing-revealed upper bound. While the 70B-ft shows moderate improvements over the base model 70B-base in tasks like Negation (33.95\% to 67.60\%) and Inference (69.47\% to 71.80\%), our method achieves performance much closer to the probing ceiling. The probing approach establishes the upper bound at 83.87\% average accuracy, with particularly strong performance in Recognition (93.10\%), HW (86.97\%), and Inference (81.59\%).
Notably, our method's performance closely tracks the probing-revealed potential across all subtasks, demonstrating effective utilization of the model's internal knowledge. For instance, in Negation and Counting tasks where 70B-base struggles (below 40\% accuracy), our approach achieves results within 10\% of the probing upper bound, significantly outperforming standard fine-tuning. 

These results demonstrate that our episodic grounding framework effectively bridges the gap between standard training approaches and the model's full potential. Notably, our method achieves this through next-token prediction, maintaining the flexibility to handle open-ended tasks like plan generation and complex reasoning. This suggests that our framework not only unlocks the model's latent capabilities but also preserves its versatility for diverse downstream applications.

\subsection{Layer-wise Analysis of Episodic Learning} 
\begin{figure}
\centering
\includegraphics[width=0.475\textwidth]{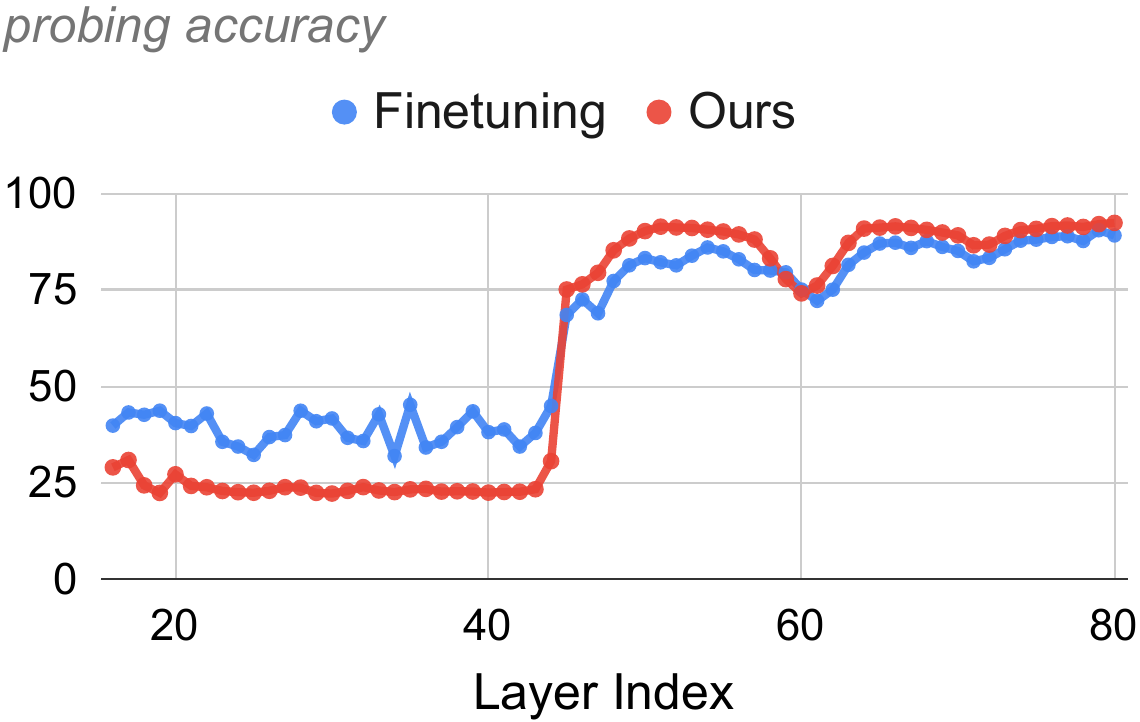}
\caption{\textcolor{black}{Layer-wise probing accuracy on Llama2-70B. Early layers (0–40) in the post-trained LM exhibit lower accuracy compared to fine-tuned models, suggesting weaker initial task alignment. In contrast, later layers (45–80) show substantial accuracy improvements, peaking at approximately 90\%. This pattern indicates enhanced alignment of task-specific representations toward output layers following weak-to-strong preference learning via MCTS.}}
\vspace{-0.12in}
\label{fig:layer-id-probing}
\end{figure}
With the success of probing inner representations in multi-choice question answering, we attempted to generalize this approach to plan generation tasks. 
However, long-form generation tasks pose a significant challenge for direct probing. Unlike multi-choice classification, where predictions can be made from static hidden representations, plan generation requires end-to-end token prediction through a continuous decoding process. This fundamental difference hinders the transferability of probing results to long-form generation. To address this, our method extends beyond direct fine-tuning by incorporating weak-to-strong DPO, aligning the model’s behavior with task-specific preferences and styles derived from episodic experience data. DPO ensures that the model’s internal capabilities are projected into coherent behaviors across diverse downstream tasks, such as plan generation, by refining the model’s preference alignment. We diagnose the impact of our method on the model’s internal representations compared to naive fine-tuning through a layer-wise probing experiment. Unlike prior approaches that aggregate hidden states from all layers to predict multi-choice answers~\cite{orgad2025llms, chetelat2025innerthoughts}, we probe each individual layer separately to assess its contribution to prediction accuracy. Figure~\ref{fig:layer-id-probing} reveals a distinct performance pattern. In early layers (Layers 0–20), the representations from our post-trained LM perform significantly worse than the fine-tuned LM, with accuracy as low as {20\%} compared to the fine-tuned model’s {40\%} on average—a drop of nearly {50\%}. This suggests that early layers in the post-trained model are less aligned with the task-relevant features. In contrast, in later layers (Layers 35–70), the post-trained LM substantially outperforms the fine-tuned model, with accuracy peaking at {90\%} in the final layers compared to {80\%} for the fine-tuned model. This accuracy boost demonstrates that our pipeline effectively pushes the model’s internal representations closer to the output, enhancing task-specific alignment and reducing the need for extensive post-processing. As shown in Table~\ref{tab:main-results}, our model (70B-ours) achieves an average accuracy of {78.93\%}, surpassing the fine-tuned model (70B-ft) at {77.96\%}, even without additional probing-based post-processing.

\subsection{Scaling Resilience in Complex Planning} 
\label{app:sec:transfer-exp}
\begin{figure}
\centering
\includegraphics[width=0.475\textwidth]{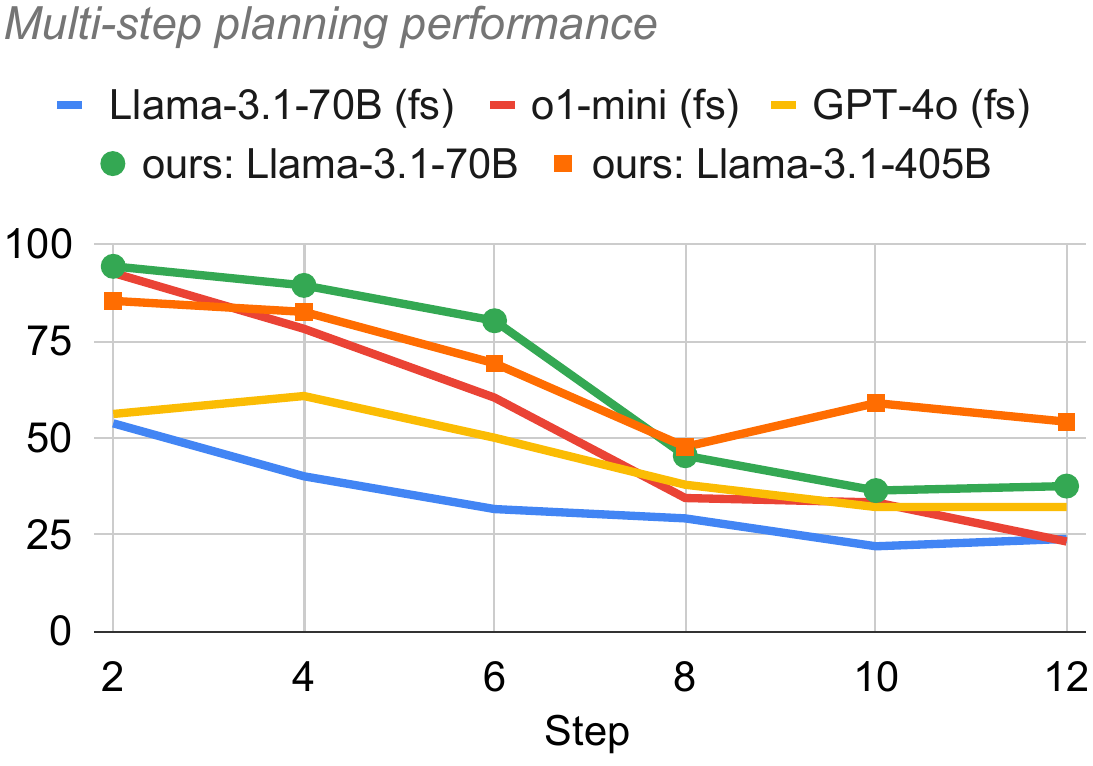}
\caption{\textcolor{black}{Planning accuracy versus the number of steps for various Los on episodic grounding tasks. Few-shot (``fs’’) methods degrade rapidly as planning complexity increases, particularly in smaller-capacity models (even when augmented with inference-time scaling methods like o1-mini). In contrast, our weak-to-strong method (applied to Llama3.1-70B and 405B) maintains high accuracy over longer planning sequences, demonstrating superior scalability and generalization capabilities from the scaled episodic grounding.}}
\vspace{-0.12in}
\label{fig:increasing-planning-complexity}
\end{figure}
\label{app:sec:transfer-exp} Figure~\ref{fig:increasing-planning-complexity} examines the performance of various LMs on planning tasks in text data sampled from a physical simulator, which are then generalized to several transferred scenarios (details in App.~\ref{app:sec:new-dataset}). Test samples are ranked by planning complexity, measured by the number of steps required to complete each task. This setup evaluates the models’ fundamental episodic grounding capabilities for planning. While inference-time scaling methods—such as few-shot reasoning and o1-mini systems—offer incremental improvements, they fail to maintain accuracy as task complexity increases. As shown in Figure~\ref{fig:increasing-planning-complexity}, few-shot performance drops rapidly after four reasoning steps. For example, accuracy for Llama3.1-70B and o1-mini declines from over {60\%} at step 2 to below {30\%} by step 8. In contrast, Llama3.1-405B sustains high accuracy, maintaining over {70\%} even at step 12, demonstrating its superior generalization capabilities. These scalability challenges arise from two fundamental factors. \textit{First}, a {capability boundary} exists, limiting the number of effective reasoning steps regardless of the method used. Previous studies~\cite{chen2024unlocking, zhang-etal-2024-working, yelongmamba} show that Chain-of-Thought reasoning and o1-like scaling methods plateau as task complexity grows, leading to diminishing returns. \textit{Second}, planning with episodic grounding requires extensive integration of social and world knowledge, which correlates strongly with model scale. Smaller LMs struggle to generalize due to their limited pretraining capacity to encode and utilize diverse contextual knowledge~\cite{zhang-etal-2024-working, yu2024kola, sun-etal-2024-head, gao2024interpretable, yuan2025superficial, zhang2025pretrained, zhang2025overcoming}. Our method (\textit{ours with Llama3.1-70B}) mitigates some of these challenges by combining episodic grounding with behavior alignment through weak-to-strong DPO. As shown in Figure~\ref{fig:increasing-planning-complexity}, it maintains higher accuracy than baseline few-shot Llama3.1-70B, sustaining over {70\%} accuracy up to step 8. These findings suggest that injecting episodic experience alone is insufficient to improve planning at scale. Instead, sustained performance requires LMs with expansive generalization capabilities, such as Llama3.1-405B, that can effectively combine episodic knowledge with broader world knowledge.

\section{Conclusion}
This work introduces a weak-to-strong framework that effectively unlocks the episodic grounding potential in LMs. Through extensive empirical validation, we show that our approach enables efficient transfer of episodic knowledge from small to large LMs, achieves peak performance in physical planning tasks, and maintains performance in complex long-step planning sequences where baseline approaches degrade. Our layered analysis and exploratory experiments suggest that our framework successfully pushes episodic knowledge processing towards the output layers of the model, while maintaining the flexibility required for open-ended planning tasks. These advances provide a practical framework for developing more capable AI systems that can effectively learn from and apply past experience in complex environments by exploiting the scaled capabilities of large LMs.

\section*{Limitations}
The current study is primarily examined on data collected from the virtual simulator, and it may not fully capture the complexity of the real-world physical interactions. Future work could explore extending it into real-world interactive robot environments.
\bibliography{ref.bib}
\appendix
\section{Predictor Module Architecture}  
\label{sec:predictor_architecture}
We provide details on the architecture of the predictor module used for the multi-choice question-answering task. The predictor is designed to map the hidden states of the LM from all layers at the last temporal position to a prediction over the possible answers. The predictor takes as input tensors of shape $(L, d)$, where $L$ is the number of layers in the LM, and $d$ is the dimension of the hidden states. The architecture consists of three main blocks:

\begin{itemize}
    \item \textbf{Block 1 (Dimension Reduction):}  
    This block begins with a normalization layer (such as LayerNorm or RMSNorm), followed by a linear transformation that reduces the dimension from $d$ to $n_1$. The output of this layer is passed through an activation function such as ReLU or Swish.

    \item \textbf{Block 2 (Further Reduction):}  
    The second block applies another normalization layer, followed by a linear transformation that reduces the dimension from $n_1$ to $n_2$. An activation function is applied to the output.

    \item \textbf{Block 3 (Prediction Layer):}  
    The output from the second block is flattened into a single vector of size $n_1 \times n_2$. A final block consisting of a normalization layer, a linear transformation, and a softmax activation function maps this vector to a probability distribution over $C$ possible answers, where $C$ is the number of answer choices in the multi-choice task.
\end{itemize}
This design ensures that the predictor can efficiently aggregate information from all layers of the LM, capturing the full representational capacity of the model. By learning a mapping from layer-wise hidden states to task-specific labels, the predictor unlocks hidden potential in the LM, leading to improved performance on the multi-choice question-answering task, as shown in Figure~\ref{fig:70B-radar-qa} and Table~\ref{tab:inner-thought-results}.

\begin{table}[t]
\centering
\resizebox{0.5\textwidth}{!}{
\begin{tabular}{lccccccc}
\toprule
\multirow{2}{*}{Config} & \multicolumn{7}{c}{Question Answering (multi-choice)} \\ 
\cmidrule(lr){2-8}
  & HW & Neg. & Recog. & Inf. & Count. & Loc. & Avg. \\  
\midrule
70B-base    & 85.82 & 33.95 & 92.35 & 69.47 & 71.65 & 80.50 & 72.29 \\
70B-ft      & 85.80 & 67.60 & 91.44 & 71.80 & 71.09 & 80.00 & \underline{77.96} \\
70B-ours    & 84.34 & 72.53 & 91.40 & 72.90 & 71.10 & 81.32 & \underline{78.93} \\
70B-probing & \textbf{86.97} & \textbf{73.65} & \textbf{93.10} & \textbf{81.59} & \textbf{81.63} & \textbf{86.30} & \textbf{83.87} \\
\bottomrule
\end{tabular}}
\caption{Llama2 results for question answering tasks with multi-choice setting. Probing means the hidden states of intermediate layers in the last position (on the direct fine-tuned LM) are sent to a trained predictor module for the multi-choice prediction.}
\label{tab:inner-thought-results}
\end{table}

\section{Thematic Scenario Data for VirtualHome Planning Task Transfer}
\label{app:sec:new-dataset}
\begin{figure*}[t]
\centering
\includegraphics[width=0.95\textwidth]{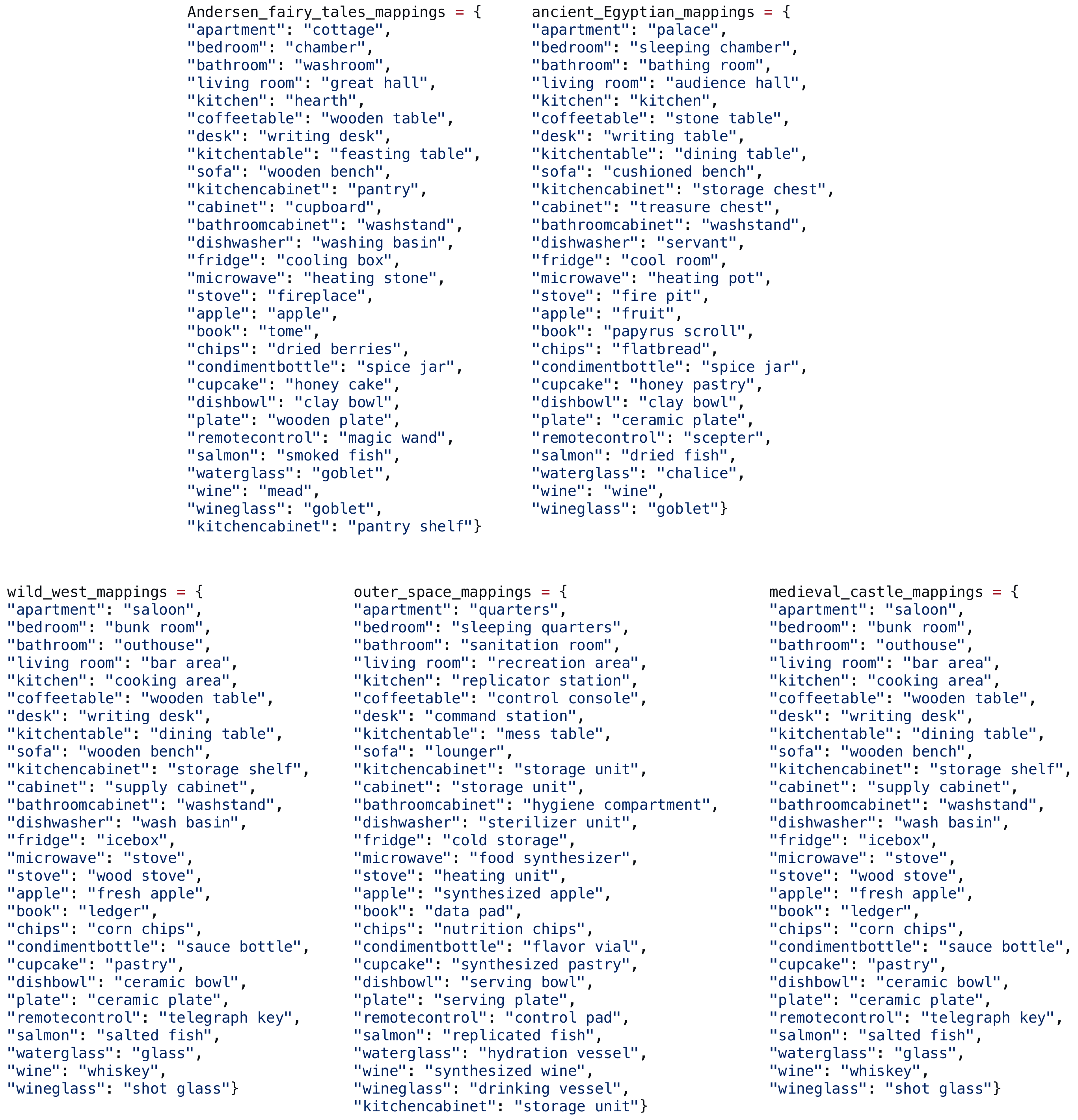}
\caption{Primary changes between the original apartment scenario and the five transferred thematic environments used in VirtualHome simulator-sampled experience data.}
\label{fig:app:transfer-mapping}
\end{figure*}
To evaluate the generalizability of our method, we introduce five new thematic scenarios: Andersen Fairy Tales, Ancient Egyptian, Wild West, Outer Space, and Medieval Castle. These environments are distinct from the original apartment setting and are not seen during the post-training phase, presenting unique challenges and contextual shifts. For example, the Medieval Castle scenario replaces modern objects like "sofa" and "microwave" with thematic equivalents such as "cushioned bench" and "heating pot," while maintaining functional consistency within the VirtualHome simulator.

Figure~\ref{fig:app:transfer-mapping} provides a visual summary of these key differences, statistically extracted and mapped to illustrate the transformation of concepts and environments across themes. These mappings ensure a \textit{self-contained and self-consistent} solution for transferring planning tasks, as the VirtualHome simulator operates as a closed-world environment. By systematically mapping objects and spaces (e.g., "apartment" to "palace," "bedroom" to "sleeping chamber"), we preserve the functional relationships and logical consistency required for effective task transfer. This approach allows us to evaluate the model's ability to adapt to dynamic environments while maintaining coherence and usability within the simulator's constraints.

\end{document}